\documentclass{bmvc2k}

\title{Where is the disease? Semi-supervised pseudo-normality synthesis from an abnormal image}

\addauthor{Yuanqi Du}{ydu6@gmu.edu}{1}
\addauthor{Quan Quan}{quan.quan@miracle.ict.ac.cn}{2}
\addauthor{Hu Han}{hanhu@ict.ac.cn}{2}
\addauthor{S. Kevin Zhou}{s.kevin.zhou@gmail.com}{2}
\addinstitution{
 George Mason University
}
\addinstitution{
 Institute of Computing Technology, Chinese Academy of Sciences
}

\runninghead{Du et al.}{Where is the disease?}

\begin{document}

\maketitle

\begin{abstract}
Pseudo-normality synthesis, which computationally generates a pseudo-normal image from an abnormal one (e.g., with lesions), is critical in many perspectives, from lesion detection, data augmentation to clinical surgery suggestion. However, it is challenging to generate high-quality pseudo-normal images in the absence of the lesion information. Thus, expensive lesion segmentation data have been introduced to provide lesion information for the generative models and improve the quality of the synthetic images. In this paper, we aim to alleviate the need of a large amount of lesion segmentation data when generating pseudo-normal images. We propose a \textbf{S}emi-supervised \textbf{M}edical \textbf{I}mage generative \textbf{LE}arning network (SMILE) which not only utilizes limited medical images with segmentation masks, but also leverages massive medical images without segmentation masks to generate realistic pseudo-normal images. Specifically, our framework consists of four primary components, including a pseudo-normality generator, a lesion segmentor, a confidence enhancer and a pseudo-abnormality reconstructor. We adopt the idea of adversarial training between the generator and the segmentor, and between the reconstructor and the segmentor to produce realistic pseudo-normal and pseudo-abnormal images. Furthermore, to reduce the number of required lesion segmentation data, we propose a confidence enhancer, which maximizes the confidence score of the segmentation prediction and enables our framework to learn in semi-supervised learning. To stabilize the model training, a step-by-step adversarial training scheme is also introduced. Extensive experiments show that our model outperforms the best state-of-the-art model by up to 6\% for data augmentation task and 3\% in generating high-quality images. Moreover, the proposed semi-supervised learning achieves comparable medical image synthesis quality with supervised learning model, using only 50\% of segmentation data.
\end{abstract}

\section{Introduction}
\label{sec:intro}

With the rapid development of deep learning and its successful applications in computer vision, natural language processing, scientific discovery, etc \cite{deng2009imagenet,devlin2018bert,graves2013speech}, more and more researchers attempt to assist the clinical diagnosis with the deep learning algorithms in many tasks, e.g. segmentation \cite{andermatt2018pathology,bowles2017brain,milletari2016v,sun2020adversarial,ye2013modality}, lesion detection \cite{alex2017generative,chen2018unsupervised,tsunoda2014pseudo} and so on \cite{shen2017deep}. Despite promising results in some tasks, there are still several main challenges left unsolved. \textbf{(1)} Most of the deep learning models are highly data-driven which require a large amount of labeled data to achieve satisfying performance. However, the labels are especially expensive for medical image data because it takes plenty of time for experts to carefully annotate the data \cite{liu2021deep}. \textbf{(2)} Deep learning algorithms often serve as a black box and are hard to assist with doctors' diagnosis. 

\begin{figure}[htb]
    \includegraphics[width=\textwidth]{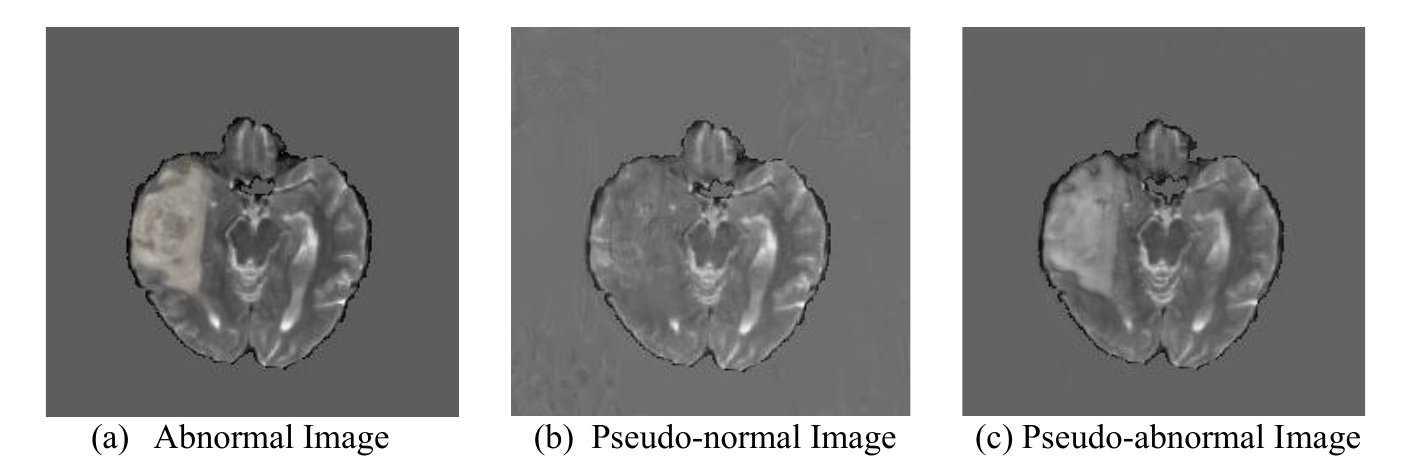}
    \caption{Visualization of abnormal, pseudo-normal, pseudo-abnormal images.} 
    \label{fig:mot}
\end{figure}

Recently, many researchers have found synthetic images generated by deep generative models, especially generative adversarial networks (GANs), can greatly improve performance of many tasks in both natural images and medical images \cite{perez2017effectiveness,goodfellow2020generative,han2018gan}. At the same time, the high-quality synthetic images generated by deep generative models can assist with doctors' diagnosis. For example, doctors can localize the lesions by comparing abnormal images with synthetic pseudo-normal images. Thus, generating medical images with great quantity and quality has become a necessary and urgent research topic.  



The idea of pseudo-normality synthesis is to generate pseudo-normal images (i.e. without lesion) from real abnormal images (i.e. with lesion). The generated pseudo-normal images are important in two aspects. \textbf{(1)} The synthetic pseudo-normal images can assist doctors with diagnosis by comparing the abnormal and pseudo-normal images \cite{tsunoda2014pseudo}. \textbf{(2)} The synthetic pseudo-normal images can be utilized as a data augmentation technique to improve performance of various downstream tasks, (e.g. lesion segmentation, lesion detection, etc.) \cite{alex2017generative,andermatt2018pathology,bowles2017brain,chen2018unsupervised,milletari2016v,sun2020adversarial,tsunoda2014pseudo,ye2013modality}. However, in most of the cases, the paired abnormal and normal images are unavailable for training such a generative network to synthesize pseudo-normal images from abnormal images. Thus, previous works \cite{baumgartner2018visual,sun2020adversarial,xia2020pseudo,yunlong2020generator} leverage unpaired normal and abnormal images to train generative networks (variants of GAN-based models) to generative pseudo-normal images from abnormal images. Nevertheless, they have several limitations: \textbf{(1)} low-quality of the generated pseudo-normal images in the absence of the segmentation labels, \textbf{(2)} high-cost of the required segmentation labels, \textbf{(3)} the dual generation problem (pseudo-abnormal image synthesis) is not yet explored. 

To solve the above limitations, we propose a \textbf{S}emi-supervised \textbf{M}edical \textbf{I}mage generative \textbf{LE}arning network (SMILE), which can leverage a small set of images with segmentation masks and a large set of images without segmentation masks to achieve robust pseudo-normality synthesis from abnormal ones. To be specific, we propose a confidence enhancer component which maximizes the confidence score of the predicted segmentation labels in our semi-supervised setting. For the third limitation, we incorporate a new component in our framework for pseudo-abnormal image synthesis. Thus, our model is capable of generating both pseudo-normal and pseudo-abnormal images, as shown in Fig. \ref{fig:mot}. Our contributions are summarized as follows:

\begin{itemize}
    \item We propose a semi-supervised generative modeling framework which uses confidence enhancer to leverage both labeled and unlabeled data to generate synthetic images.
    \item We propose a confidence enhancement technique which enforces maximization of the certainty over the predicted segmentation mask and is especially helpful in the absence of ground-truth segmentation labels. 
    \item Extensive experiments show that our model outperforms the best of the state-of-the-art methods by up to 3\% in generating high-quality images and 6\% for data augmentation task. When using only 50\% of labeled data, the proposed model is able to achieve comparable pseudo-normality synthesis results to fully supervised-methods.
\end{itemize}


\section{Related Works}
\label{sec:related_work}

\subsection{Pseudo-normality Synthesis}
Pseudo-normality synthesis is important in various perspectives, including clinical suggestion for doctors and data augmentation for many downstream tasks \cite{andermatt2018pathology,bowles2017brain,sun2020adversarial,ye2013modality,zhu2017unpaired}. Thanks to the satisfying visual effects of the synthetic images, generative adversarial nets have been dominating in the pseudo-normality synthesis task. In the previous work, \cite{sun2020adversarial} utilizes anomaly detection, which treats lesions as outliers, to guide the generation process. 
In a later work \cite{xia2020pseudo}, they introduce the idea of adversarial training, which co-trains a predictor to make binary predictions over the synthetic images. The predictor is expected to detect whether the synthetic images are normal (i.e. 0) or abnormal (i.e. 1). The generator is expected to generate realistic pseudo-normality images which fool the predictor to believe that the synthetic pseudo-normality images are real. Additionally, a reconstructor is also introduced to reconstruct back the pseudo-abnormality images taking the segmentation mask and the pseudo-normality images as input. 
In the most recent work \cite{yunlong2020generator}, they extend the idea of adversarial training and co-train a segmentor which provides more detailed information about the lesion than the predictor used in \cite{xia2020pseudo}.
Besides GAN-like architectures, in work \cite{chen2018unsupervised}, they adopt a variational auto-encoders (VAE) to learn the distribution of the normal images and generate pseudo-normality images based on the learnt constrained latent representations. However, they suffer from the poor reconstruction visual effects of VAE as well as hard to eliminate small lesions. 


\subsection{Deep Generative Models for Medical Image Synthesis}
To alleviate the need of the large amount of data required by deep learning models, deep generative models have been proposed firstly to generate synthetic data for effective data augmentation \cite{goodfellow2020generative}. In medical image analysis, deep generative models are also utilized for data augmentation and anonymization \cite{shin2018medical,han2019combining}. 
In the previous work, various downstream tasks have been utilized to guide the generator to produce high-quality synthetic images, such as anomaly detection, lesion detection, lesion segmentation, etc. For example, \cite{sun2020adversarial} utilizes anomaly detection, which treats lesions as anomalies, and learn a generator which removes the lesion from the original images. However, since there is no ground-truth for the pseudo-abnormal images, the labels are obtained by simply removing everything within the contour of the brains from the images. Obviously, the assumption is too strong for an abnormal image which should maintain the structures and tissues within the brains. Later work in \cite{xia2020pseudo, yunlong2020generator} eliminate the assumption by providing more information about the lesions in the images (e.g. binary masks).

\subsection{Generative Adversarial Learning in Medical Image Analysis}
Generative Adversarial Nets (GAN) \cite{goodfellow2020generative} have been proposed to generate realistic images and achieved satisfying results in natural images \cite{zhu2017unpaired}. The fundamental mechanism of the GAN-based approach is to co-train a generator and a discriminator together. The generator is expected to generate realistic images which fool the discriminator to believe the synthetic images are real, while the discriminator is expected to distinguish the real and synthetic images. Later, the idea of adversarial training which one model attempts to fool the other model by generating realistic images is largely adopted in several different forms. For example, conditional GAN \cite{mirza2014conditional} utilizes the idea of adversarial training to make the generator learn to generate images of different classes. In medical image analysis, the idea of adversarial training is also largely utilized \cite{han2018gan,yi2019generative}. Specifically, \cite{chen2017deeplab,tack2018knee,xue2018segan} have shown that adversarial training can improve the performance of various medical image analysis tasks, such as segmentation, detection, etc.

\begin{table}
    \centering
    \begin{tabular}{|c|c|} \hline
    Symbol & Description \\ \hline
    $G$ & pseudo-normality generator\\
    $S$ & lesion segmentor\\
    $R$ & pseudo-abnormality reconstructor\\
    $F$ & confidence enhancer \\
    $I$ & input image w/ segmentation label\\
    $L$ & segmentation label\\
    $P_{L}$ & generated pseudo-label\\
    $U$ & unlabeled input image\\
    $M$ & predicted segmentation mask\\
    $P_{n}$ & generated pseudo-normality image \\
    $P_{ab}$ & reconstructed pseudo-abnomality image \\
    \hline
    \end{tabular}
    \vspace{3mm}
    \caption{List of notations.}
    \label{tab:not}
\end{table}

\section{Methodology}
\label{sec:method}

The overall framework includes three main parts (1) pseudo-normality generation, (2) adversarial segmentation and (3) pseudo-abnormality generation. The confidence enhancement is for the semi-supervised learning setting. The list of notations used throughout this section is first introduced in Table \ref{tab:not}. Then, all the three parts are illustrated in the following sections.

\begin{figure*}[htb]
    \centering
    \includegraphics[width=.80\textwidth]{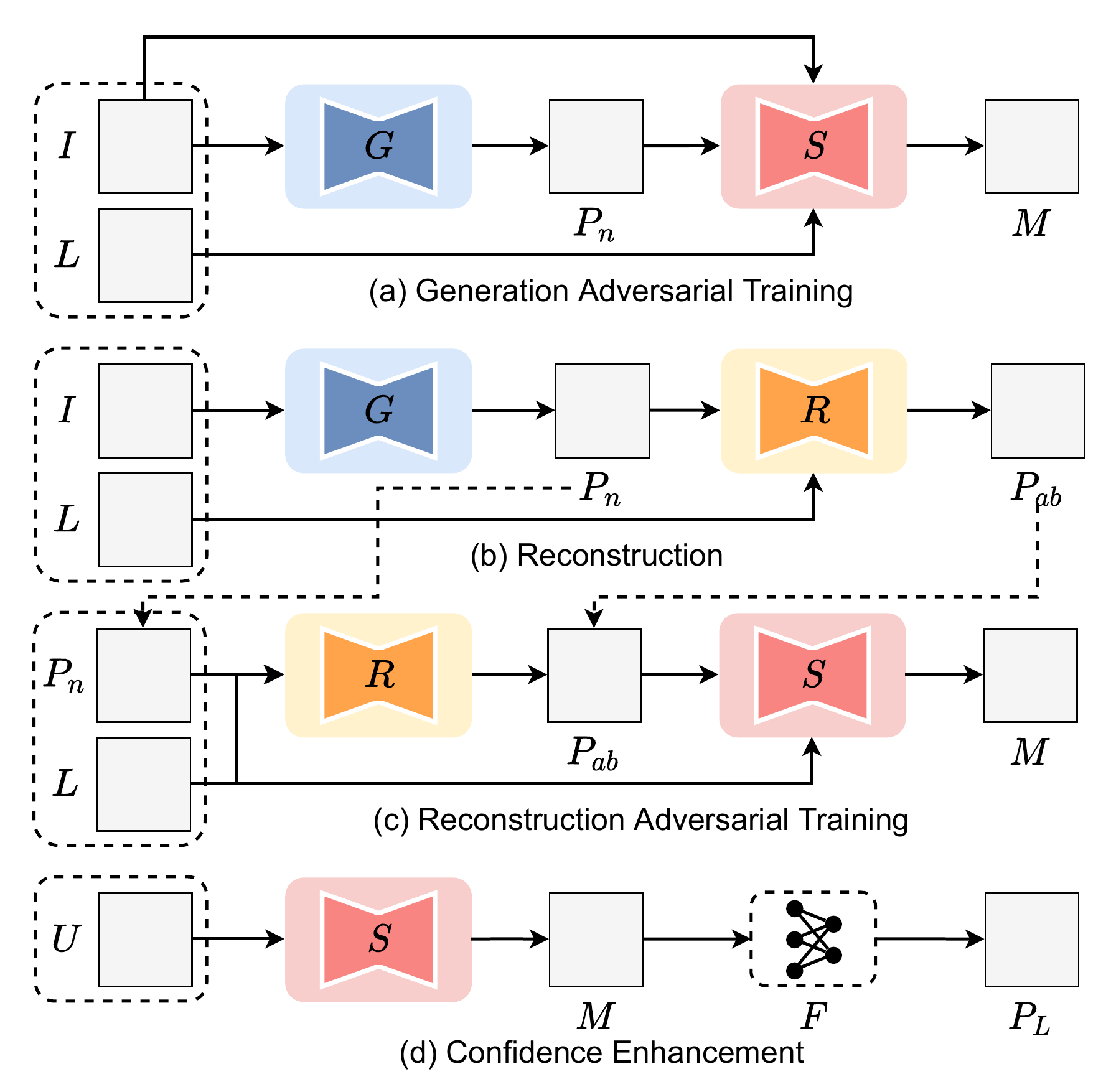}
    \caption{(a) Pseudo-normality Generation Adversarial Training (b) Pseudo-abnormality Reconstruction (c) Pseudo-abnormality Reconstruction Adversarial Training (d) Semi-supervised Confidence Enhancement}
    \label{fig:model_arc}
\end{figure*}

\subsection{Pseudo-normality Generation}
We adopt a U-Net \cite{ronneberger2015u} model, which is a widely used image segmentation architecture, to build our generator because the architecture allows different resolutions of information to flow through and skip connection to pass message along, which both are critical for the generation task. The loss function of the generator consists of two parts, the first part is a pixel-wise measurement on the quality of the generated pseudo-normality image, and the second part measures how good the generated pseudo-normality image is by the prediction from the discriminaotr. Therefore, the loss of the generator is defined as follows,
\begin{equation}
    L_{gen} = 
    L_{mse}((1-M) \odot I ,\: (1-M) \odot \mathbf{G}(I)) + L_{ce}(\mathbf{S}(\mathbf{G}(I)),\: M)
\end{equation}
where $M$ is the segmentation mask of the expected normal image (i.e. zeros matrix), $L_{mse}$ refers to a pixel-wise $L_2$ loss and $L_{ce}$ refers to the cross-entropy loss. The pixel-wise generation $L_2$ loss enforces the generator to maintain the identity of the no-lesion part. Notably, in order to keep the model consistent and aware of the presence or absence of the lesion, we also train the model to generate pseudo-normality image with the input real normal image. In this case, the model is expected to output the original image.



\subsection{Adversarial Segmentation}
In terms of the segmentor, we also adopt a U-Net architecture due to its excellent performance in segmentation task \cite{ronneberger2015u} and a softmax function in the end to produce segmentation masks. The main objective for the segmentor is to recognize the lesion behind the generated pseudo-normality images and recognize the fake-lesion in the generated pseudo-abnormality images, as:
\begin{equation}
    \begin{split}
     L_{seg} = & L_{ce}(\mathbf{S}(\mathbf{G}(I)),\;M) + L_{ce}(\mathbf{S}(P_{ab}),\;M)
    \end{split}
\end{equation}
where $M$ is the segmentation mask and $P_{ab}$ is the reconstructed pseudo-abnormality image.

\subsection{Pseudo-abnormality Generation}
To fully utilize the segmentation labels to provide more supervised signals to the model as well as enrich the dataset better during data argumentation, we introduce a reconstructor at the end of our model to learn to reconstruct the input image with the segmentation mask and the pseudo-normality image. As in the generator, we adopt a similar U-Net structure, except with a two-channel input. The objective function is to minimize the pixel-wise loss between the reconstructed image and the original input image.
\begin{equation}
    P_{ab} = \mathbf{R}(\mathbf{G}(I),\:M)\:\: L_{recons} = L_{mse}(P_{ab},\: I)
\end{equation}

\subsection{Semi-supervised Objective}
In our semi-supervised training setting, the input image is with unknown type, either normal or abnormal. The input image $U$ first goes through the segmenter. Two following things will happen: (1) the image trains the segmentor without label by enhancing the confidence of the segmentor (as in Eq. \ref{eq:4}), (2) the segmentor assigns a pseudo-label to the image. After that, the training is same as it is in the supervised setting, with the assigned pseudo-label. In this setting, the segmentor is trained to be more confident about the segmentation mask, which in turns provides a more precise pseudo-label and leads to a better solution for both the segmentor and the generator. The loss function for the confidence enhancement of the segmentor is:
\begin{equation}
    P_{L} =
    \begin{cases}
      1, & \text{if}\ M>\epsilon \\
      0, & \text{otherwise}
    \end{cases}
    \label{eq:4}
  \end{equation}
\begin{equation}
    L_{semi} = L_{ce}(S(U),P_{L})
\end{equation}
where $\epsilon$ is a desired confidence threshold, which we set as $0.9$ in our model. It is also recommended to dynamically update $\epsilon$ by the average confidence value every epoch. In this way, the confidence of the segmentor with respect to the accuracy of the predicted segmentation mask is maximized. It is especially useful under the semi-supervised learning which no ground-truth is present to guide the training of the framework.

\subsection{Training Scheme}
We adopt a step-by-step pre-training and fine-tuning training scheme as shown in Fig. \ref{fig:model_arc}, We first train $G$ and $S$ in an adversarial setting. Then, we train $R$ by fixing $S$ and $G$. Next, we train $R$ and $S$ again in an adversarial setting. Finally, we fine-tune the framework with $G$, $S$, $R$ together. We adopt a similar training scheme in the semi-supervised setting, except the segmentation mask is predicted from the segmentor $S$.

\section{Experiments}
\label{sec:exp}
\subsection{Dataset}
We consider the publicly available dataset, \textbf{Multimodal Brain Tumor Segmentation Challenge 2019 (BraTS19)} \cite{bakas2017advancing,menze2014multimodal}
to evaluate our framework.  We take the training set of the BraTS19 challenge dataset. There are 259 GBM (i.e. glioblastoma) and 76 LGG (i.e. lower-grade glioma) volumes of magnetic resonance imaging (MRI) in the dataset. Each of them is skull-stripped, interpolated to an isotropic spacing of $1mm^3$ and co-registered to the same anatomical template. In every volume, there are four different modalities (i.e. T1, T2, T1c and Flair) with 240x240 slice available. We take the T2 modality of the GBM and split train/validation/test set by 130/104/25 volumes, respectively. We clip the intensities to [0, $V_{99.5}$] of each volume, where $V_{99.5}$ accounts for top 99.5\% pixel values of the volume.



\begin{figure}[htb]
    \centering
    \includegraphics[width=\textwidth]{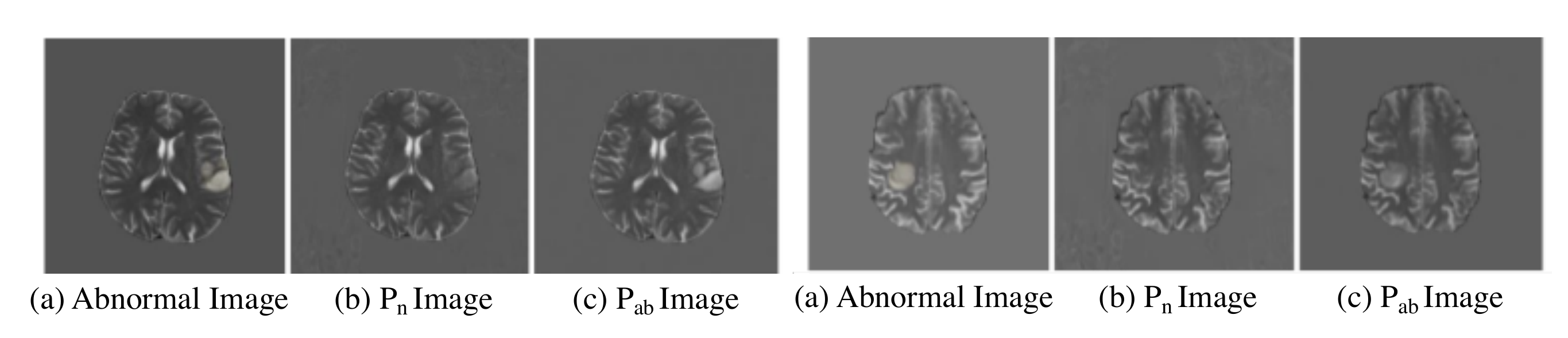}
    \caption{Visualization of the randomly selected abnormal and synthetic pseudo-normality and pseudo-abnormality images.}
    \label{fig:qual_eval}
\end{figure}

\subsection{Comparison Methods}
We consider four related frameworks for the supervised learning task of pseudo-normality synthesis. The models are introduced below. We take the baseline implementation from their open-source codes. \textbf{VA-GAN}\cite{baumgartner2018visual} uses a generator which can predict an additive maps and then generate a pseudo-normality images and a discriminator which attempts to distinguish between the generated images and the real images. Additionally, they aim to learn a least required additive map to achieve the realistic pseudo-normality image generation by minimizing a $L_1$ loss. \textbf{ANT-GAN}\cite{sun2020adversarial} is a variant of a CycleGAN which it enforces the consistency loss between the pseudo-normality and pseudo-abnormality images. Compared with our approach, they do not intend adversarial training in the network, while they only use the pixel-wise $L_2$ normality consistency loss.
\textbf{PHS-GAN}\cite{xia2020pseudo} introduces adversarial training with a discriminator to distinguish between normal and abnormal images, while \textbf{GVS}\cite{yunlong2020generator} takes segmentation label as input and co-train a discriminator to segment the lesion in the normal and abnormal images. 

\subsection{Evaluation Metrics}
For quantitative evaluation, we select two metrics proposed in \cite{xia2020pseudo}, to evaluate the healthiness and identity, in which healthiness refers to the effects of the erased tumor, while identity refers to the maintenance of the original structure. 
\textbf{Healthiness ($h$)} measures how normal the generated pseudo-normality image is. The evaluation function of $h$ is defined as:
\begin{equation}
    h = 1-\frac{E_{x_p \sim P}[N(\textbf{S}_{pred}(\textbf{G}(I))]}{E_{x_p\sim P}[N(\textbf{S}_{pred}(I))]}
\end{equation}
\textbf{Identity ($iD$)} measures the structure maintenance of the generated pseudo-normality image. The evaluation function of $iD$ is defined as:
\begin{equation}
    iD = MS-SSIM[(1-M)\odot \textbf{G}(I), (1-M)\odot I]
\end{equation}
where $MS-SSIM$ stands for multiscale structural similarity \cite{wang2003multiscale}.

\subsection{Experiment Details}
The framework is trained with an Adam Optimizer. The batch size is set as 8 and the learning rate is set as 1e-4. The model is trained 10 epochs for each phase of the training loop. All the experiments are conducted on a 64-bit machine with two NVIDIA GPUs (TITAN RTX, 1770MHz, 24GB GDDR6).

\section{Discussion}
\label{sec:discussion}

\begin{table}
    \centering
    \begin{tabular}{|c|c|c|} \hline
    Methods & identity $iD \uparrow$ & healthiness $h \uparrow$ \\ \hline
    VA-GAN\cite{baumgartner2018visual} & 0.809&0.605 \\
    PHS-GAN\cite{xia2020pseudo}&0.960 &0.731 \\
    ANT-GAN\cite{sun2020adversarial} & 0.969&0.747  \\
    GVS\cite{yunlong2020generator} &0.997& 0.792\\\hline
    SMILE-50 & 0.978 & 0.653\\\hline
    SMILE-75 &\textbf{0.987} & \textbf{0.810}\\\hline
    SMILE-sup & \textbf{1.000} & \textbf{0.822}\\\hline
    \end{tabular}
\vspace{3mm}
\caption{Quantitative evaluation on identity and healthiness metric of our proposed SMILE framework and the other baseline models.}
\label{tab}
\end{table}

\subsection{Evaluation on Generation Performance}
In Figure \ref{fig:qual_eval}, we visualize randomly selected triplets of (abnormal images, pseudo-normal images, pseudo-abnormal images). In Table \ref{tab}, we evaluate the $iD$ and $h$ evaluation metrics of our proposed model as well as the baseline models. In Table \ref{tab}, it is clear to observe that our model outperforms other models in both evaluation metrics. Notably, GVS ranks the second, even though the framework is very simple. That demonstrates the adversarial training with segmentation task is important to provide detailed information (e.g. shape, location) about the lesion and further improve the quality of the synthetic images. The performance of ANT-GAN and PHS-GAN are comparable and way better than VA-GAN, which illustrates that adversarial training could help improve the quality of the synthetic images. In summary, utilizing segmentation task as an adversarial training task could greatly improve the quality of the synthetic images. Rather, the additional pseudo-abnormality reconstructor further improve the quality of the synthetic pseudo-normal images by ~2\%. For the semi-supervised setting, we use only 75\% segmentation-labeled data, along with 25\% unlabeled data to train out network, and achieve comparable results with the fully-supervised learning setting. In practice, we do not involve any extra unlabeled data. Rather, we remove the label of the 25\% labeled data from the same training set to produce the unlabeled data, thanks to the proposed confidence maximization technique, which continues to train the model, even when the labels are missing. Besides, we evaluate how the amount of labeled data influence our semi-supervised learning performance in table \ref{tab} (a). It is worth noting that the labeled data we use is at least 50\% of the whole labeled training dataset because the framework crashes with only few labeled data and we leave it as a future study.



\subsection{Evaluation on Data Augmentation}
We evaluate the effectiveness of the method when working as a data augmentation task performance by providing synthetic pseudo-normality and pseudo-abnormality data for the downstream lesion segmentation tasks. (Note that we do not apply traditional data augmentation (e.g. rotation, crop, etc.) here.) We show the performance (Dice score) of the models in Table \ref{tab:data_aug}. To be specific, we set up the experiments by utilizing our training dataset to generate the data-augmented dataset, which we generate two types of data (i.e. pseudo-normality, pseudo-abnormality) with the same amount as in the training dataset. We compare the performance of each different type of augmented-dataset, namely, without augmentation (None), with pseudo-normality only (PN or GVS \cite{yunlong2020generator}), with pseudo-abnormality only (PA) and with both pseudo-normality and -abnormality (Both). Note, the setting PN is the baseline model, GVS \cite{yunlong2020generator}, which produces the best performance over the current state-of-the-art models. For each of the data-augmented experiment, adding pseudo-images means incorporating the synthetic data into the training data. The result shows that both the generated pseudo-normality and pseudo-abnormality data can improve the performance of the downstream task (i.e. segmentation) when used for data augmentation.

\begin{table}
    \centering
    \begin{tabular}{|c|c|} \hline
    Methods & $Dice \uparrow$ \\ \hline
    None & 0.72 \\\hline
    Pseudo-normality (GVS)&0.76  \\\hline
    Pseudo-abnormality & 0.77  \\\hline
    Both &\textbf{0.82}\\\hline
    \end{tabular}
\vspace{3mm}
\caption{Quantitative evaluation on dice score of the downstream lesion segmentation task.}
\label{tab:data_aug}
\end{table}

\section{Conclusion}
\label{sec:conclusion}
In this paper, we proposed a \textbf{S}emi-supervised \textbf{M}edical \textbf{I}mage generative \textbf{LE}aning (SMILE) framework to generate pseudo-normality and pseudo-abnormality images to assist in medical image analysis tasks through eye screening and data augmentation. A confidence enhancement technique is introduced for semi-supervised generative learning. Extensive experiment results suggest that our proposed SMILE model can generate images with better quality and support better data augmentation than the state-of-the-art models. We plan to study more data-efficient ways (e.g. self-supervised learning) for generative learning in medical images in the future.

\bibliography{WID}
\end{document}